# CLASH WRIST - A hardware to increase the capability of CLASH fruit gripper to use environment constraints exploration

Werner Friedl and Máximo A. Roa

*Abstract*— Humans use environmental constraints (EC) in manipulation to compensate for uncertainties in their world model. The same principle was recently applied to robotics, so that soft underactuated hands improve their grasping capability by using environmental constraints exploitation (ECE) [1]. Due to orientation of the robotic hand for example in the EC wall grasp, the length of the robot wrist plus the hand length gets quite important, if objects are grasp out of a box [2] . Most of the modern cobots have quite long wrist, so we have constructed a two degree of freedom wrist for the CLASH [3], to solve this problem (Fig. 1).

## I. INTRODUCTION

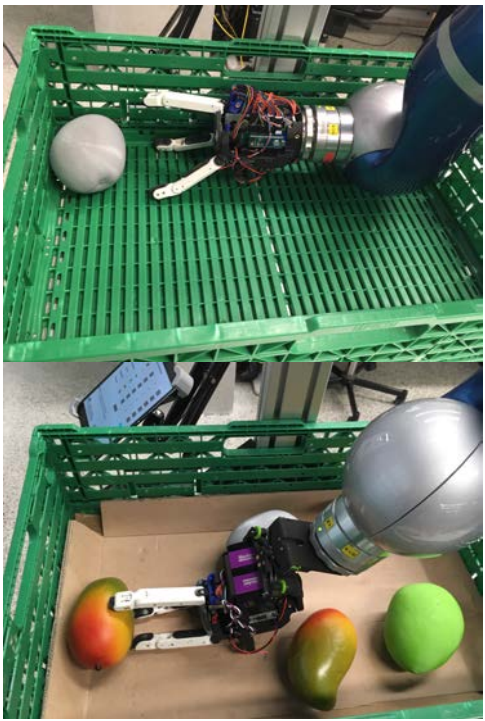

Fig. 1: Without Wrist Wallgrasp works only for quite empty IFCO on one wall - with wrist on all walls and in crowed scenarios

If we look at the human anatomy, the distance between wrist flexion and fingertip is only around 178 mm. Furthermore, the width of the human wrist is also smaller than the hand width. This allows an effective hand workspace. A compare of these dimensions for cobots with mounted CLASH and CLASH with wrist on LWA3 is show in table I.

*This work has been funded by the European Commission's eighth Framework Program as part of the project SoMa (grant H2020-ICT-645599).
All authors are with the Institute of Robotics and Mechatronics, German Aerospace Center (DLR), Wessling, Germany. {firstname.lastname}@dlr.de

| Cobot | length with CLASH | width |
|---|---|---|
| DLR LWA 3 | 335 mm | 166 mm |
| DLR CLASH Wrist | 230 mm | 80 mm |
| Franka emica | 420 mm | mm 120 mm |
| KUKA IIWA | 480 mm | 142 mm |
| KUKA IISY | 405 mm | – mm |

TABLE I: Dimensions for cobots with mounted CLASH

In principle, an adapter with a fixed twist of the hand to the last joint of the robot can help to bring the robot wrist out of box. This idea was developed by UNIPI/ITT and used in OCADO use case during the EU project SOMA (Fig. 2). Still the top grasp is more important for grasping objects out of a box and the adapter decreases the workspace of the robot, because the robot can not use the last joint to rotate the hand directly. Instead the rotation of the hand has to come from several joints. To check the influence of the adapter we calculated the capability map [4] for the pre grasp positions of apples in the box. The inverse kinematic gives the results for 1152 positions in the box and for every position the possible solutions for rotations in ten degree steps of the hand. As you can see in Fig. 3 the workspace of the robot with mounted adapter is much worser then without. An extra actuated wrist allowed to use the normal top grasp capability of the robot but also increase the workspace in the blue areas. Thanks to the use of rapid prototyping technology and standard components, the CLASH wrist is a cheap platform for logistic grasping experimentation.

## II. DESIGN

The wrist offers two extra active degrees of freedom (DOF) for positioning the hand with respect to the object. (see Fig. 4). The first axis (palmar flexion and dorsal extension) is mainly for orienting the hand parallel to the box surface to improve the capability for a wall grasps. A second scenario for this DOF is the reorientation of the hand in a shelf to reach the object behind the first row. The second axis (abduction and adduction) allowed to grasp smaller object from the side with a two finger grasp vertical to the surface. Both DOFs are actuated by coupled DITEX servos. For the first DOF the motors drive in opposite direction. For the second degree of freedom the motors drive in the same direction and move the tendons. The blue tendons are fixed to two sliders (see Fig. 5), which are connecting the two parts of the rolling joint. The flexion spoon is rolling on part B and keeps it position by the four black pretension tendons. This mechanism has a ratio of two and allow the output to move with the double angle then the sliders. Overall the ratio between servo motion and output is 1/3 due to the ratio of 1/6 between servo pulley diameter and slider pulley. Furthermore the mechanism is saving length instead to a conventional rotation joint. The wrist actuators are connected to the CLASH pcb and are controlled by the hand interface.

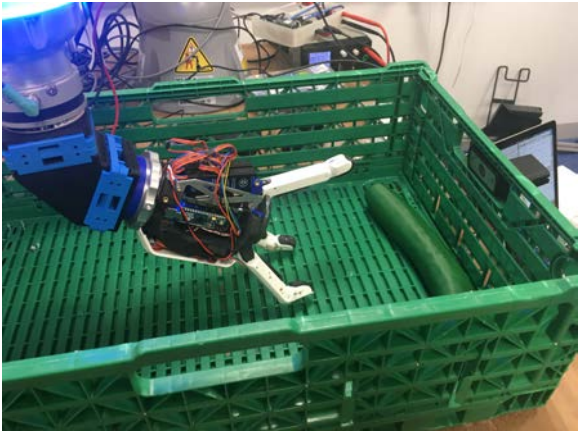

Fig. 2: CLASH mounted with 60 degree rotation adapter on KUKA IIWA

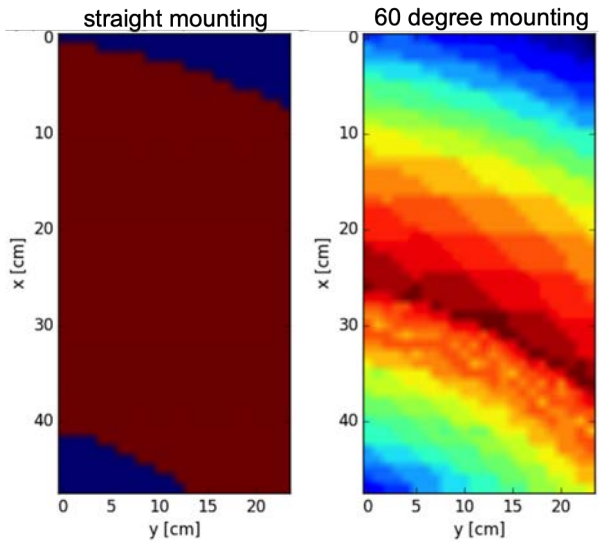

Fig. 3: topgrasp capability map - dark red: 100 precent of orientations, blue: zero percent

| Parameter | max torque | range of motion | maximum velocity |
|---|---|---|---|
| dorsal flexion | 5.2 Nm | 100 degree | 500 degree/s |
| Abduction | 15.6 Nm | +/- 40 degree | 166 degree/s |

TABLE II: Datasheet of CLASH wrist

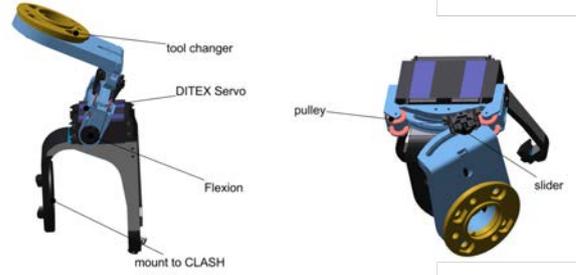

Fig. 4: Left: maximum flexion, right: maximum abbduction

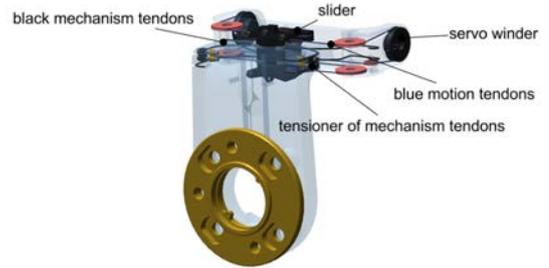

Fig. 5: tendons of wrist

## III. EXPERIMENT

The wrist was tested in a crowed box of mangos and had to grasp one lying at a wall with a EC wall grasp. Fig . 6 shows the experiment, the arm can hold its tool center point orientations during the grasp. The necessary hand orientation are done by the wrist. If we compare the grasping result out of the system benchmark [5] for example mangos , the wrist leads to 33 percent better performance as a fixed twist of the hand and 13 percent less planning errors.

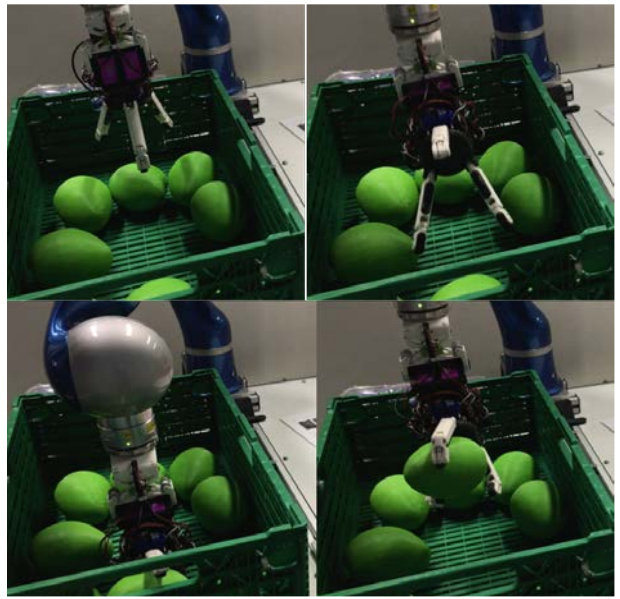

Fig. 6: wall grasp sequence of a mango

Another interesting scenario is grasping objects out of shelfs. As example we use the ACRV picking benchmark [6] scenario to show, that CLASH with wrist can grasp object behind other objects in the shelf (see Fig: 7). The robot and wrist positions were taught, due to an integration of the wrist in the planner is missing in the moment.

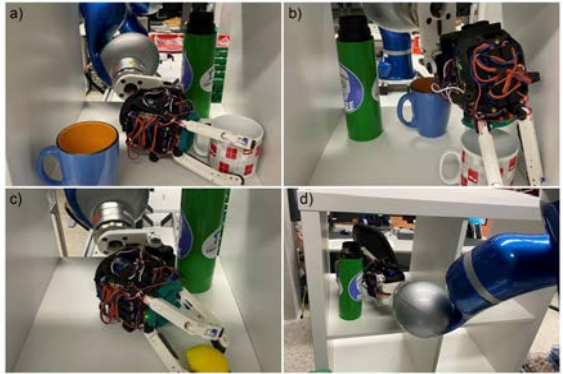

Fig. 7: Grasping objects out of an IKEA shelf (ACRV Benchmark)

## IV. CONCLUSION AND FUTURE WORK

The introduced wrist design is also interesting for humanoid robot arm due to is great workspace and small dimensions compared to a kardan joint. Furthermore for the CLASH the wrist should be integrated in the planner to use it more efficient.

## REFERENCES

[1] C. Eppner, R. Deimel, J. Alvarez-Ruiz, M. Maertens, and O. Brock, "Exploitation of Environmental Constraints in Human and Robotic Grasping," *Int. J. Robotics Research*, vol. 34, no. 7, pp. 1021–1038, 2015.
[2] F. Negrello, S. Mghames, G. Grioli, M. Catalano, M. Garabini, and A. Bicchi, "A Compact Soft Parallel Wrist for Grasping in Narrow Spaces," *IEEE Robotics and Automation Letters*, p. submitted, 2019.
[3] W. Friedl, H. Höppner, F. Schmidt, M. A. Roa, and M. Grebenstein, "CLASH: Compliant low cost antagonistic servo hands," in *IEEE/RSJ Int Conf. Intelligent Robots and Systems (IROS)*, 2018.
[4] F. Zacharias, C. Borst, and G. Hirzinger, "Online generation of reachable grasps for dexterous manipulation using a representation of the reachable workspace," in *2009 International Conference on Advanced Robotics*, June 2009, pp. 1–8.
[5] P. Triantafyllou, H. Mnyusiwalla, P. Sotiropoulos, M. Roa, D. Russell, and G. Deacon, "A benchmarking framework for systematic evaluation of robotic pick-and-place systems in an industrial grocery setting," 06 2019.
[6] J. Leitner, A. W. Tow, N. Sünderhauf, J. E. Dean, J. W. Durham, M. Cooper, M. Eich, C. Lehnert, R. Mangels, C. McCool, P. T. Kujala, L. Nicholson, T. Pham, J. Sergeant, L. Wu, F. Zhang, B. Upcroft, and P. Corke, "The acrv picking benchmark: A robotic shelf picking benchmark to foster reproducible research," in *2017 IEEE International Conference on Robotics and Automation (ICRA)*, May 2017, pp. 4705–4712.